\documentclass[11pt,a4paper]{article}
\usepackage{amsmath,graphicx,multirow,adjustbox,fullpage,url}
\usepackage[rgb]{xcolor}
\usepackage{flushend}
\usepackage[font=small,skip=3pt]{caption}
\usepackage{enumitem}
\usepackage{cite}
\usepackage{gensymb}

\date{}
\title{ICIP 2022 Challenge on Parasitic Egg Detection and Classification in Microscopic Images: Dataset, Methods and Results}

\author{ Nantheera Anantrasirichai$^1$,  Thanarat H. Chalidabhongse$^2$, \\ Duangdao Palasuwan$^3$,  Korranat Naruenatthanaset$^2$, Thananop Kobchaisawat$^2$,\\ Nuntiporn Nunthanasup$^3$, Kanyarat Boonpeng$^3$, \\  Xudong Ma$^1$ and Alin Achim$^1$\footnote{This work was supported by the Newton Fund Institutional Links, British Council [grant number 623714323]; the National Research Council of Thailand [grant number N31A640268]; the Chulalongkorn University Technology Center's Innovation Fund.} \\ $^1$Visual Information Laboratory, University of Bristol, UK, \\
	$^2$Department of Computer Engineering,  Chulalongkorn University, Thailand, \\
	$^3$Oxidation in Red Cell Disorders Research Unit, \\ Department of Clinical Microscopy, Chulalongkorn University, Thailand}
\begin{document}
%
\maketitle
\begin{abstract}
Manual examination of faecal smear samples to identify the existence of parasitic eggs is very time-consuming and can only be done by specialists. Therefore, an automated system is required to tackle this problem since it can relate to serious intestinal parasitic infections.
This paper reviews the ICIP 2022 Challenge on parasitic egg detection and classification in microscopic images. We describe a new dataset for this application, which is the largest dataset of its kind. The methods used by participants in the challenge are summarised and discussed along with their results.  
\end{abstract}
\section{Introduction}
\label{sec:intro}

Intestinal parasitic infections remain among the leading causes of morbidity worldwide, especially in tropical and sub-tropical areas with high temperate climates. According to WHO, approximately 1.5 billion people, or 24\% of the world’s population, were infected with soil-transmitted helminth infections (STH), and more than 800 million children worldwide required preventive chemotherapy for STH in 2020\footnote{\url{https://www.who.int/news-room/fact-sheets/detail/soil-transmitted-helminth-infections}}. Most infections can cause diarrhea and other symptoms, such as malnutrition and anaemia, particularly in children, who may suffer from growth failure. Most infected persons can also shed cysts or eggs in their living environment, and unwittingly cause transmission of parasites to other individuals. An improvement in personal hygiene, better sanitation and a widespread health education campaign have reduced helminthic infections, but protozoa are still present even in asymptomatic hosts leading to chronic diseases. In developing countries, intestinal protozoa and STH have been recognised as one of the most significant causes of illness. Diagnosis of intestinal parasites is usually based on direct examination in the laboratory. Unfortunately, this time-consuming method (30 min/sample) shows low sensitivity and requires an experienced and skilled medical laboratory technologist, being thus impractical for on-site use. This means an automate routine faecal examination for parasitic diseases is essential.

\begin{table}[]
    \centering
    \caption{Size and resolution ranges of 11 parasitic egg types in our dataset}
    \begin{tabular}{lcc}
    \hline
      Type & size ($\mu$m) & width (pixels)  \\
    \hline
      Ascaris lumbricoides  & 60$\times$85 & 131-439 \\
      Capillaria philippinensis  & 20-22$\times$36-45 & 56-234 \\
      Enterobius vermicularis & 20-30$\times$50-60 & 76-272 \\
      Fasciolopsis buski & 80-85$\times$130-140 & 170-806 \\
      Hookworm egg & 36-40$\times$64-76 & 114-410 \\
      Hymenolepis diminuta & 60-80 & 132-461\\
      Hymenolepis nana & 30-47 & 95-300 \\
      Opisthorchis viverrine & 11-12$\times$22-32 & 41-186 \\
      Paragonimus spp & 77-80 & 116-477\\
      Taenia spp. egg & 30-35 & 85-244 \\
      Trichuris trichiura & 22-23$\times$50-54 & 76-395 \\
    \hline
    \end{tabular}
    \label{tab:typesandsizes}
\end{table}

This challenge aims to encourage and highlight novel strategies with a focus on robustness and accuracy in data-driven technologies to automatically detect parasitic eggs and also identify the egg type in compound microscopy images. We provide a new large dataset containing 11 parasitic egg types. The dataset is well-balanced between classes, with varying sizes from very small, 22$\times$11~$\mu$m  (Opisthorchis viverrine), to very large, 140$\times$85~$\mu$m  (Fasciolopsis buski), as shown in Table \ref{tab:typesandsizes}.
Our dataset exhibits real and synthetic degradation due to out-of-focus, noise, motion blur, colour, and light imbalance. A wide variety of image quality and appearances create a large range of different image characteristics. This aims to enhance the robustness of the detection models. 

The challenge has gathered experts in the fields of image processing, medical imaging and computer vision, which should not be limited to the domain of microscopic imaging as knowledge transfer can benefit the growth of the community. Our dataset has been hitherto downloaded from more than 20 countries in 4 continents. The outcome of the challenge could further improve and assist diagnosis in real clinical use, and even automate detection and identification of intestinal parasite eggs, which can be used by non-experts.

\section{Related work}
\label{sec:related}

\noindent \textbf{Existing datasets}: All available microscopic datasets for parasitic egg detection and classification are generally small in both numbers of categories and images.
For example, the Parasitology Laboratory of Kansas State University provides 5 types of parasitic eggs with only a total of 33 images ({https://www.k-state.edu/parasitology/}). The Japan Advanced Institute of Science and Technology offers 8 types with 213 images in total \cite{Viet2019}. A 4-class dataset of 162 images captured with 10$\times$ magnification is reported in \cite{suwannaphong2021parasiticmsc}. The larger dataset in \cite{Quinn2016} containing 3,327 1000$\times$ magnified microscopic images has only three parasitic egg types.

\vspace{3mm}
\noindent \textbf{Existing detection and classification methods}: Traditional methods  of parasite egg detection involve pre-processing, feature extraction and classification with support vector machine (SVM) \cite{Ray2019,Avci2009} or artificial neural networks (ANN) \cite{Yang2001,Ray2019}. The performances are however limited because the methods cannot deal with complex datasets.
Deep learning has recently shown its ability and high efficiency to tackle problems in object detection \cite{Anantrasirichai:AI:2021}. The state-of-the-art methods of parasite egg detection and classification are therefore based on this technology, especially the convolutional neural networks (CNNs). The early work employs seven convolutional layers  trained from scratch \cite{Quinn2016}. Later, a transfer learning strategy using a pretrained AlexNet \cite{Krizhevsky2017} and a pretrained ResNet50 \cite{He2015} was employed to classify parasitic egg types, and the positions of the eggs were identified using sliding window technique \cite{suwannaphong2021parasiticmsc}. Some methods follow semantic segmentation using UNet architecture \cite{Li2019} or a fully convolutional network (FCN) \cite{Najgebauer:Microscopic:2019}. These techniques however require full pixel-wise annotation. More modern object detection approaches based on a Faster-RCNN \cite{Ren:FasterRCNN:2015} are employed in \cite{Viet2019} and \cite{Mayo:Detection:2022}, achieving promising results. 

\section{Dataset for the Challenge}
\label{sec:dataset}

Our dataset, called \textbf{Chula-ParasiteEgg-11}, contains 11 types of parasitic eggs from faecal smear samples. With a total of 13,750 microscopic images, it has become the largest dataset of its kind. In this challenge, the training and testing datasets are randomly divided resulting 1,000 images/class for training and approximately 250 images/class for testing. Some testing images contain two or more types of parasitic eggs and some testing images do not have any eggs. The labels take the form of bounding boxes. 
The visual quality of some images is degraded to replicate the real scenario, in which the images are acquired with speed or automated system in different lighting conditions.

\begin{figure}
    \centering
    \includegraphics[width=0.5\columnwidth]{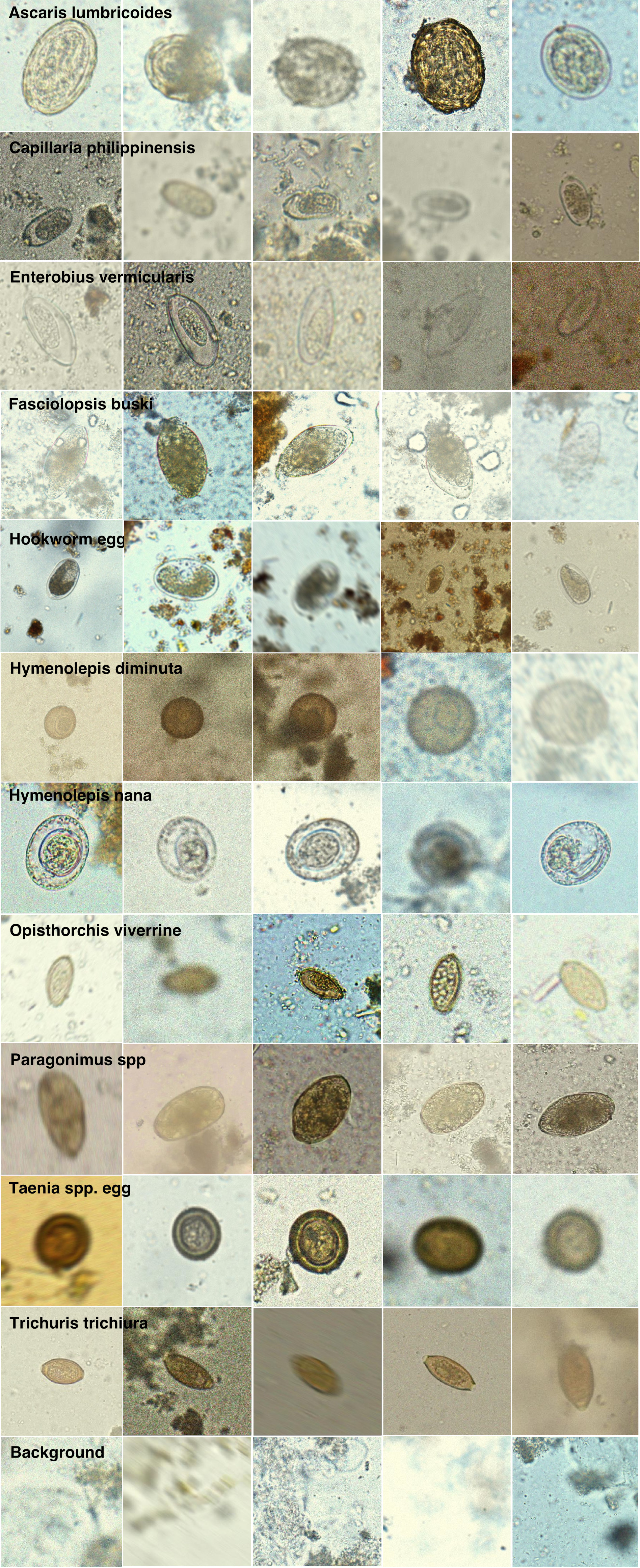}
    \caption{Examples of 11 parasitic egg images after quality degradation. Note that these cropped images are not true scaling as for better visualisation. The last row shows variation of the background and debris.}
    \label{fig:dataset}
\end{figure}

Samples from the dataset are shown in Fig. \ref{fig:dataset}. Their average size ranges approximately in the interval of 15-140 $\mu$m. The details of the size and diameter of each parasitic egg type are listed in Table \ref{tab:typesandsizes}. 
Training and testing datasets are available on the IEEE Dataport \cite{Palasuwan:parasitic:2022}.

\vspace{3mm}
\noindent \textbf{Parasite specimens and acquisition}: The stool samples of parasitic eggs used for the experiments in this study were prepared in laboratory conditions at Department of Clinical Microscopy, Faculty of Allied Health Sciences, Chulalongkorn University. The specimens were fixed with 10\% formalin prior to storing at 4\degree C. Stool samples were examined microscopically by direct simple smear. After the samples were prepared, they were immediately examined visually using a light microscope with 40$\times$ magnification. Three researchers checked the egg types and the majority vote was used as the label. The bounding boxes indicating the position of the egg were the average amongst the majority.

A total of 1,250 images of each parasitic egg type was acquired from several different devices, including Canon EOS 70D camera body with Olympus BX53 microscopes, DS-Fi2 Nikon camera body with Nikon Eclipse Ni microscopes, Samsung Galaxy J7 Prime phone, iPhone 12 and 13 with either 10$\times$ times eyepieces lens of Nikon Eclipse Ni or Olympus BX53 devices. These result in different resolutions, different lighting and setting conditions.

\vspace{3mm}
\noindent \textbf{Image degradation}: The visual quality of the microscopic images captured above are degraded so that i) the detection competition is more challenging, ii) the robustness of the detection models is enhanced, and iii) the possible images captured by motorised stage microscope are replicated. The quality degradation procedure is as follows.
\begin{itemize} [leftmargin=*]
\setlength\itemsep{-0.2em}
    \item Crop each side by a random value from 0 to 30\%,
\item  Apply Gaussian blur with a standard deviation between  0.0 - 3.0,
\item  Apply motion blur (10\% of total images) with a random kernel size of 3- 35 pixels and orientation of 0-360 degree,
\item  Apply  Gaussian noise with a standard deviation between 0.0 - 25.5 (to each colour channel separately),
\item  Apply  Poisson noise with a lambda (the expected rate of occurrences) between 0.0 - 5.0,
\item   Adjust image saturation by adding a value in a range of -25 - +25 to the S channel of the HSV colour space,
\item   Adjust contrast using gamma correction with a gamma value between 0.5 - 2.0.
\end{itemize}

After the challenge competition, we also made the original-quality images available for research in fields such as image restoration, image enhancement or super-resolution.

\section{Methods}
\label{sec:methods}

All methods in this competition exploited deep learning techniques. Most of them were developed based on the state of the art in object detection, including YOLOv5 \cite{He:Report:2022, Pratama:PARASITIC:2022}, Fast-RCNN \cite{Mayo:Detection:2022, Tureckova:high:2022}, EfficientDet \cite{AlDahoul:local:2022},
Cascade R-CNN \cite{He:Report:2022, Pang:Report:2022, Lu:Report:2022}, CBNetV2 \cite{NEGU:Report:2022, Muralidhara:Improved:2022},  
 CenterNet2 \cite{Tango:Background:2022}, Task-aligned One-stage object Detection (TOOD) \cite{Aung:high:2022, Tureckova:high:2022}, and RetinaNet \cite{Pho:attention:2022,Hongcharoen:Report:2022}. These methods are convolutional neural networks with various backbone architectures, where the most popular architecture is based on ResNet blocks \cite{Aung:high:2022, Tureckova:high:2022, Tango:Background:2022, Muralidhara:Improved:2022, Santaquiteria:Parasitic:2022,Pho:attention:2022, Bandara:Rethinking:2022}. 
 Most methods employed Feature Pyramid Network (FPN) to extract multi-scale features \cite{Aung:high:2022,AlDahoul:local:2022, Pho:attention:2022, Muralidhara:Improved:2022}, whilst some methods employed Transformer-based architectures as feature extractors \cite{Pang:Report:2022, Pedraza:Parasitic:2022, Tsai:Report:2022, Lu:Report:2022, Hongcharoen:Report:2022}. 
The models were pretrained with COCO dataset \cite{Aung:high:2022, Tango:Background:2022,NEGU:Report:2022, Muralidhara:Improved:2022, Santaquiteria:Parasitic:2022, Pedraza:Parasitic:2022, Lu:Report:2022,Tsai:Report:2022,Pho:attention:2022}, or  ImageNet-1K \cite{Pang:Report:2022}. These methods have proved that using pre-trained weights significantly improves the detection performance.
The final models of some teams were fused from multiple models obtained by cross validation and fine-tuning parameters \cite{NEGU:Report:2022,Tureckova:high:2022,Santaquiteria:Parasitic:2022, Aung:high:2022}. Further improvement has been achieved with small object detection mechanisms \cite{Tureckova:high:2022,Muralidhara:Improved:2022} and pseudo labelling strategy \cite{NEGU:Report:2022, Aung:high:2022}.


Team NEGU \cite{NEGU:Report:2022} achieved the best performance. Their framework is based on CBNetV2 \cite{Liang:CBNetV2:2021}, which is composite FPNs. Three models were trained and the weighted boxes fusion was employed for the ensemble. The runner-up is Team ZUSTF4 \cite{He:Report:2022}. Their framework integrates YOLOv5 and Cascade R-CNN, trained with standard data augmentation. In third place is Team Bad Crew \cite{Aung:high:2022}, where the results comprise five models: cascade-RCNN, generalized focal loss with FPN, HTC HRNetV2p, HTC x101, and TOOD. The teams in first and third place improved their models with pseudo labelling strategies. The results shown in Fig. \ref{fig:F1vsIoU} and Table \ref{tab:results} however were reproduced directly from their submitted models without retraining with the hidden data.

\section{Results}
\label{sec:results}

\subsection{Evaluation criteria} 
All training images were released along with 2,200 unknown-labelled testing images. The participants uploaded their results of the testing dataset to the website, where a Mean Intersection-over-Union (mIoU) was the main criteria for evaluation. Each type of parasitic eggs was assessed and assigned an Intersection-over-Union (IoU) score, then the average of their IoU scores of all parasitic egg types were calculated, finally resulting in the mIoU score. Submissions were ranked according to the criteria and shown on the leaderboard. The top-performing teams were asked to submit their code and models for final checking and tested on 550 hidden testing images for the final validation. We also computed the F1-score (IoU$\geq$0.5) for each egg type to evaluate the performance using the precision and recall. The mAP@[0.5:0.95] shows the average of mAP (mean Average Precision) over different IoU thresholds, ranging from 0.5 to 0.95, step 0.05.  If a further comparison is needed for tie-breaking, the computational speed will be used as the last criterion.


\begin{figure}
    \centering
    \includegraphics[width=0.6\columnwidth]{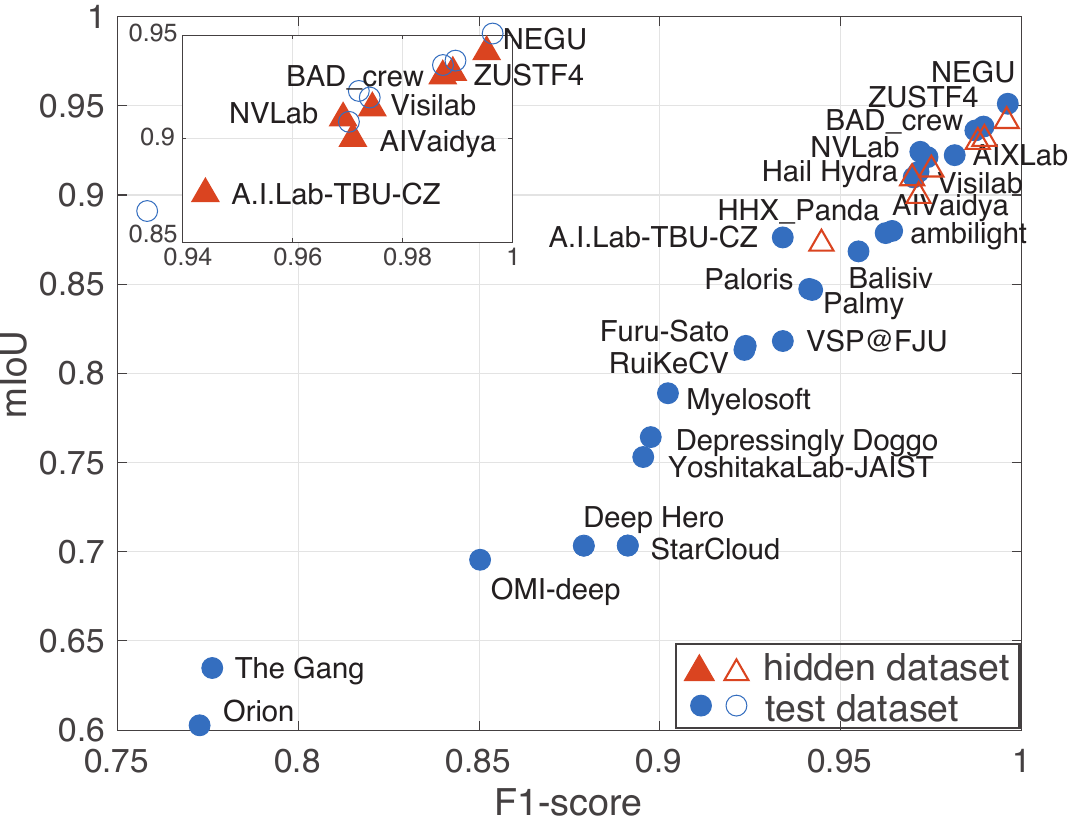}
    \includegraphics[width=0.6\columnwidth]{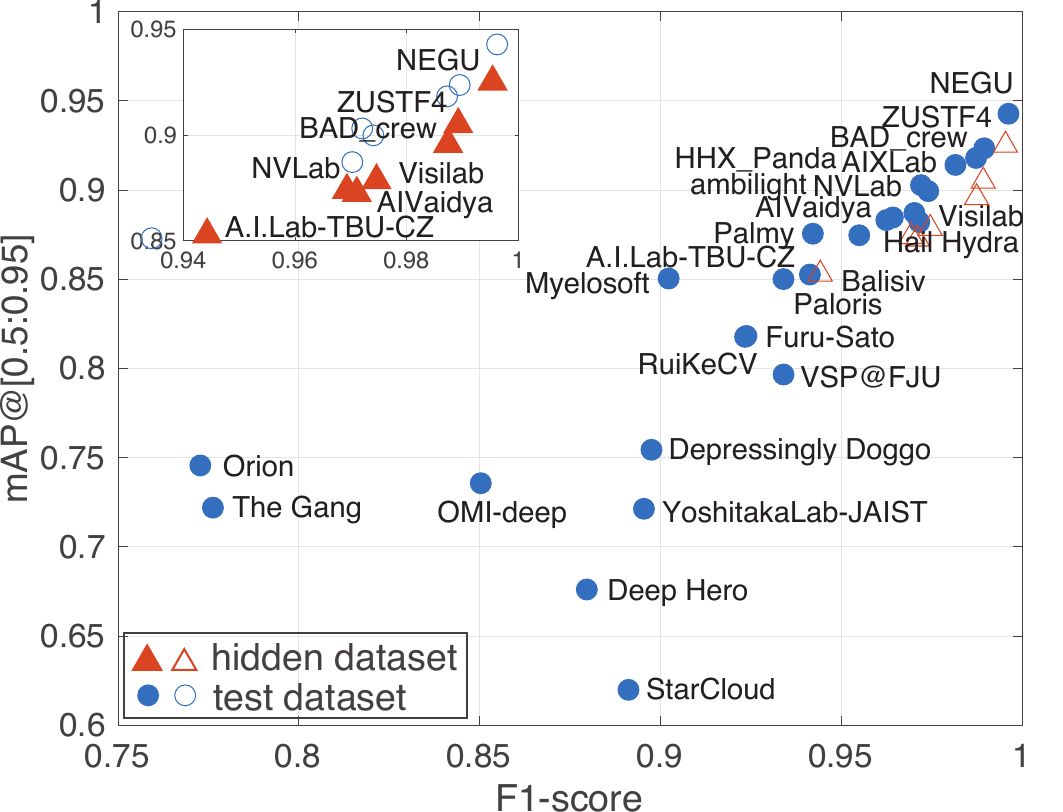}
    \caption{(Top) Combined F1-score and mIoU values. (Bottom) Combined F1-score and mAP values. Blue circles are the results from the 2200-test set we gave to participants. Red triangles are the results from the 550-hidden set.}
    \label{fig:F1vsIoU}
    \label{fig:F1vsmAP}
\end{figure}
\begin{table}[]
    \centering
    \caption{Average top-5 detection results for each parasitic egg type (FNR=false negative rate)}
    \footnotesize
    \begin{tabular}{lccccc}
    \hline
      Type & mIoU & F1score & Precision & Recall & FNR(\%)\\
    \hline
     A.L.  & 0.82 & 0.90 & 0.94 & 0.87 & 11.96 \\
C.P.  & 0.77 & 0.90 & 0.86 & 0.95 & 3.94 \\
E.V. & 0.81 & 0.92 & 0.89 & 0.96 & 3.25 \\
F.B. & 0.89 & 0.97 & 0.97 & 0.98 & 2.26 \\
H.E. & 0.92 & 0.99 & 0.98 & 0.99 & 0.61 \\
H.D. & 0.82 & 0.91 & 0.85 & 1.00 & 0.25 \\
H.N. & 0.83 & 0.93 & 0.94 & 0.92 & 7.30 \\
O.V. & 0.81 & 0.93 & 0.89 & 0.97 & 2.50 \\
P.S. & 0.80 & 0.90 & 0.86 & 0.97 & 2.27 \\
T.S. & 0.93 & 0.98 & 0.97 & 1.00 & 0.25 \\
T.T. & 0.90 & 0.97 & 0.98 & 0.96 & 4.25 \\
    \hline
    \end{tabular}
    \label{tab:results}
\end{table}

\vspace{-2mm}
\subsection{Result comparison} 
There were 82 teams registered but only 30 teams submitted their results. Fig. \ref{fig:F1vsIoU} show the mIoU vs. F1-score and the mAP vs. F1-score of each team. The winner, NEGU, achieved a mIoU of 0.942 and F1-score of 0.995 on the hidden data. The two runner-ups, ZUSTF4 and BAD crew, achieved close performance, where the mIoUs are  0.932 and 0.930, and F1-score are 0.989 and 0.987, respectively. The mAPs of these top three are 0.925, 0.905 and 0.896. However, the performances of individual class detection are quite different, particularly on the Capillaria philippinensis eggs, where NEGU and ZUSTF4 teams achieve a mIoU of 0.86 and 0.69, respectively. Fig. \ref{fig:F1vsIoU} bottom reveals that some teams achieve better mAPs but worse F1-scores than other teams, such as Orion and The Gang. This is because they achieve high precision and low recall, implying high fault negatives (miss detection).
Testing with the hidden data, the average performances of the top teams in term of mIoU, F1-score and mAP dropped by 0.82\%, 0.02\% and 1.88\%, respectively, compared to testing with the disclosed test data.

Table \ref{tab:results} shows the average performance of the  top-five methods on detecting each type of parasitic eggs. The Capillaria philippinensis egg is the most difficult to detect -- lowest values of mIoU and F1-score. The detected boxes are usually larger than the actual ground truth. The large eggs, i.e. Ascaris lumbricoides and Hymenolepis nana, are missed easier than other types (high false negative rate (FNR or miss rate)). This occurs when the high level of blur deteriorates detail and texture inside the eggs, leading to misclassification between them. The Taenia spp. eggs are the easiest to detect, even though their size is not large. This is because their shape and texture are different from debris in the background.

\section{Conclusions}
\label{sec:conclusion}
This paper reviews the ICIP 2022 Challenge on parasitic egg detection and classification in microscopic images on the data and the methods used. A new dataset comprising 11 parasitic egg types was captured with several microscopes and lenses, generating a total of 13,750 images. The quality of each image was randomly degraded with noise, Gaussian blur, motion blur and different levels of brightness, and colour adjustment. All methods used in this competition exploited deep learning technologies, where the architectures were based on state of the art object detection methods. Through this challenge, the performance of parasitic egg detection and classification reaches 0.942, 0.995 and 0.925 in terms of mIoU, F1-score and mAP, respectively.




\small
\bibliographystyle{IEEEbib}
\bibliography{challenge}

\end{document}